\documentclass{article}

\usepackage{PRIMEarxiv}

\usepackage[utf8]{inputenc} 
\usepackage[T1]{fontenc}    
\usepackage{hyperref}       
\usepackage{url}            
\usepackage{booktabs}       
\usepackage{amsfonts}       
\usepackage{nicefrac}       
\usepackage{microtype}      
\usepackage{lipsum}
\usepackage{fancyhdr}       
\usepackage{graphicx}       
\usepackage{algorithm}
\usepackage{algpseudocode}
\usepackage{listings}
\usepackage{xcolor}

\definecolor{codegreen}{rgb}{0,0.6,0}
\definecolor{codegray}{rgb}{0.5,0.5,0.5}
\definecolor{codepurple}{rgb}{0.58,0,0.82}
\definecolor{backcolour}{rgb}{0.95,0.95,0.92}

\lstdefinestyle{mystyle}{
    backgroundcolor=\color{backcolour},   
    commentstyle=\color{codegreen},
    keywordstyle=\color{magenta},
    numberstyle=\tiny\color{codegray},
    stringstyle=\color{codepurple},
    basicstyle=\ttfamily\footnotesize,
    breakatwhitespace=false,         
    breaklines=true,                 
    captionpos=b,                    
    keepspaces=true,                 
    numbers=left,                    
    numbersep=5pt,                  
    showspaces=false,                
    showstringspaces=false,
    showtabs=false,                  
    tabsize=2
}

\usepackage{amsmath}
\graphicspath{{media/}}     

\usepackage{lipsum}

\title{
Naeural AI OS - Decentralized ubiquitous computing MLOps execution engine 
}

\author{
   C. Bleoțiu, Ș. Saraev, B. Hobeanu, \& A.I. Damian \\\\\\
}

\begin{document}
\maketitle

\begin{abstract}
Over the past few years, ubiquitous, or pervasive computing has gained popularity as the primary approach for a wide range of applications, including enterprise-grade systems, consumer applications, and gaming systems. Ubiquitous computing refers to the integration of computing technologies into everyday objects and environments, creating a network of interconnected devices that can communicate with each other and with humans. By using ubiquitous computing technologies, communities can become more connected and efficient, with members able to communicate and collaborate more easily. This enabled interconnectedness and collaboration can lead to a more successful and sustainable community.
The spread of ubiquitous computing, however, has emphasized the importance of automated learning and smart applications in general. Even though there have been significant strides in Artificial Intelligence and Deep Learning, large scale adoption has been hesitant due to mounting pressure on expensive and highly complex cloud numerical-compute infrastructures. Adopting, and even developing, practical machine learning systems can come with prohibitive costs, not only in terms of complex infrastructures but also of solid expertise in Data Science and Machine Learning.
In this paper we present an innovative approach\footnote{This project incorporates open-source components developed with the support of financing grant SMIS 156084 "ReDeN - Decentralized Network for Neural Processing", provided by the Romanian Competitiveness Operational Programme. We extend our gratitude for this support, which has been instrumental in advancing our work and enabling us to share these resources with the international community.} for low-code development and deployment of end-to-end AI cooperative application pipelines. We address infrastructure allocation, costs, and secure job distribution in a fully decentralized global cooperative community based on tokenized economics.
\end{abstract}

\section{Introduction}

\subsection{From Web 1.0 to 2.0 and finally to 3.0}

Pervasive computing applications have been a part of our lives in various forms for quite some time - from smartwatches to smartphones, from smart homes to smart traffic lights. Although some argue that ubiquitous or pervasive computing is equivalent to decentralization \cite{hansmann2003pervasive} \cite{conti2012looking} \cite{hansmann2013pervasive}, most systems still depend on intricate proprietary cloud systems. The concept of fog computing \cite{chan2017fog} involves the cloud infrastructure handling the heavy lifting of hardware and logic, while edge devices manage IoT data acquisition, thin client user interaction, and basic edge processing.

Besides this, if we examine the Internet in terms of evolution, we can suggest that as the transition from \emph{Web 1.0}, which was read-only, to \emph{Web 2.0} introduced a massive content explosion via creators, influencers, and advocates, the  logical and natural progression would be for \emph{Web 3.0} to enable transparency and democratization. The current practice and theory have shown that decentralization  \cite{voshmgir2019token} \cite{sunyaev2021token}, can bring significant benefits to human society. Blockchain-based technologies \cite{nofer2017blockchain} \cite{zheng2018blockchain} \cite{monrat2019survey} have the potential to democratize data, enable peer-to-peer compute sharing, provide affordable go-to-market options for content creators and app developers. These technologies offer secure, transparent, trustless, and low-cost micro-transactions.

Artificial Intelligence (AI), machine (deep) learning in particular, is the most important catalyst for the development of intelligent applications and pervasive computing. In the era of \emph{ChatGPT}, people acknowledge the transformative power of Artificial Intelligence, but its adoption is slow due to the high costs associated with developing and running AI-enabled applications.

The primary focus in the last decade has been on developing and deploying cloud-based AI applications, largely due to the efforts of major players, such as \emph{FAANG} \cite{pisal2021rise}, to somehow make high-performance GPUs accessible to the broader audience. Despite the efforts made, there are several problems that remain unresolved or only partially resolved, which hinder the widespread adoption of AI applications and systems. These issues arise from the high cost of AI expertise and the extensive range of skills required for developing and deploying production-grade AI systems. Additionally, the significant expenses and energy consumption of GPU computing infrastructure contribute to substantial carbon footprints. To overcome these challenges, designing and implementing deep neural models for practical use cases requires a combination of scientific and business expertise, along with practical experience. 

All in all, it appears that the creation of AI applications is still far from being a truly democratized process, and several problems must be addressed before it becomes a horizontal approach that is widely accessible to content and service creators.

\subsection{The Ratio1 Protocol vision}

We are convinced that a true democratization of Artificial Intelligence would lower the costs of development and deployment of smart AI services while keeping the costs of processing at a minimum. Our vision is two-fold: (i) to convert computing devices to real assets and (ii) to democratize AI. 
We strongly believe that society will benefit immensely from transforming a wide range of computing devices, such as laptops, gaming stations, cloud infrastructure, consoles, and crypto-currency mining rigs, into real assets. These devices can then generate active income for their owners by executing jobs that have real societal and industrial value in a decentralized, peer-to-peer distributed fashion. Our AI democratization vision aims to bring AI use cases to end-consumers by reducing the go-to-market time, resources, and costs required by service providers, content creators, and software developers.

However, democratizing AI is not just a matter of technological efficiency; it is a \textbf{fundamental shift towards accessibility, affordability, and inclusion} in AI-driven innovation. Today, AI advancements are often concentrated in the hands of large corporations with the resources to develop, deploy, and commercialize cutting-edge models. This centralized model creates barriers to entry, limiting opportunities for individuals, small businesses, and underserved communities to participate in the AI economy.

Our approach aims to \textbf{eliminate these barriers} by creating an open, trustless ecosystem where AI processing power is distributed across a decentralized network. This brings several key societal benefits:

\begin{itemize}
    \item \textbf{Accessibility}: By leveraging distributed computing, AI services can reach users who lack access to high-performance infrastructure. Whether it's a student in a rural area needing access to AI-powered learning tools or a healthcare provider in a developing region requiring AI diagnostics, our vision ensures that advanced AI applications are no longer a privilege of the few.
    
    \item \textbf{Affordability}: Decentralized AI deployment reduces reliance on expensive, proprietary cloud solutions. By optimizing resource sharing and allowing anyone with computing power to contribute, we enable cost-effective AI services that are affordable for startups, independent developers, and organizations with limited budgets.
    
    \item \textbf{Inclusion in Underserved Markets}: Many industries and regions suffer from a lack of AI integration due to prohibitive costs or infrastructure constraints. Our model fosters AI innovation in emerging markets by lowering the financial and technological barriers to adoption. This can lead to breakthroughs in sectors such as agriculture, education, and healthcare, where AI has the potential to drive significant positive change.
\end{itemize}

Our paper outlines an architectural approach to enable the democratization of AI, which aims to reduce the carbon footprints of complex systems and create ecosystems where AI application deployment relies on trustless sharing of processing power. The proposed approach also intends to reduce the time gap between the incubation of an idea and the launch of a marketable product. Our proposal builds upon existing research and development in the field of Machine Learning Operations, particularly upon the \emph{SOLIS}\cite{ciobanu2021solis} framework architecture. The additions include: decentralized job distribution layer, improved message security with internal blockchain consensus, and integration with EVM \cite{wood2014ethereum} compatible networks on-chain.

\subsection{Summarizing our progress so far}

In our previous work \cite{ciobanu2021solis}, we proposed an end-to-end architecture and methodology aimed at standardizing critical stages of production-grade machine learning pipelines, with a particular focus on edge-based systems. The primary goal was to offer technological independence, freedom to operate, and versatility in composing and deploying business rules. This resulted in an end-to-end MLOps framework that supports multiple jobs and workers, tensor framework agnosticism, and pre-built templates and serving pipelines for prominent frameworks such as \emph{TensorFlow} \cite{abadi2016tensorflow} and \emph{PyTorch} \cite{paszke2019pytorch}. 
The currently proposed architecture enables seamless integration with a wide array of data sources, including CCTV cameras, relational databases, flat files, and sensor networks, while facilitating rapid deployment of business rules through low-code and no-code approaches. The component which lies at the foundation of this architecture, the \textbf{Naeural Execution Engine}, can be easily configured for IoT-based protocols like \emph{MQTT} \cite{hunkeler2008mqtt} and \emph{AMQP} \cite{amqp}, as well as web \emph{REST} endpoints, with other endpoints implementable via a low-code plugin approach.

Building on the vision and objectives of our previous work, additional research and development have been carried out in the field of trustless job distribution within decentralized networks. Various peer-to-peer job distribution mechanisms and templates inspired by \emph{MapReduce}\cite{dean2008mapreduce} patterns have been devised. However, unlike traditional \emph{MapReduce} approaches, a significant paradigm shift was necessary due to the trustless nature of the decentralized peer-to-peer network and the heterogeneity of the worker pool's processing capacity and availability.

\section{Related work}

\subsection{MLOps, DevOps and productization of AI}
To the best of our knowledge, there has been no significant advancements in the field of Machine Learning Operations (MLOps) and Development Operations (DevOps) since our initial publication \cite{ciobanu2021solis}. The focus of our research was to reduce the barriers to entry for the development of end-to-end AI-enabled systems or large-scale IoT-based applications. Classic DevOps methodologies are insufficient for the deployment of machine learning-based products due to the high computational demands. Therefore, it is necessary to optimize the use of available resources while maintaining a stable environment. Typically, machine learning applications require the use of both CPU and GPU, where the former is used for heuristics and business logic, while the latter is used for inference and data preprocessing/postprocessing. As a result, a specialized solution is required.

One such solution is \emph{Kubeflow} \cite{bisong2019kubeflow}, a framework-agnostic, open-source tool designed to streamline machine learning pipelines on \emph{Kubernetes}. It supports most stages of the machine learning lifecycle, including data preprocessing but without data acquisition of any sort, model training and retraining, prediction serving and versioning, or service management. The main drawback of \emph{Kubeflow} is that it requires extensive prior knowledge of Data Science and \emph{Kubernetes}. Another drawback, that we are aware of, is that \emph{Kubeflow} does not support full end-to-end application pipeline development but is rather focused only on the pure MLOps.

Another option is \emph{TensorFlow Serving} \cite{olston2017tensorflow}, which allows for easy deployment of \emph{TensorFlow}-based machine learning models through a low-code solution via a \emph{gRPC} or \emph{HTTP} endpoint. However, it does not provide tools for training, data acquisition, or business logic - not to mention highly complex end-to-end pipelines required in various types of applications.

\emph{ONNX} \cite{bai2019onnx} is an open format for representing machine learning models, regardless of the technologies used to build them. It currently supports conversions from popular machine learning frameworks such as \emph{TensorFlow}\cite{abadi2016tensorflow}, \emph{PyTorch}\cite{paszke2019pytorch}, and \emph{SciKit-Learn}, and its purpose is to provide a framework-agnostic interface for machine learning applications.

Finally, in relation to the \textbf{Naeural} architecture, while the above-mentioned frameworks and technologies could be considered competing approaches for MLOps, most of them are in fact technologies that could be easily integrated and encapsulated within the \textbf{Naeural Execution Engine Pipelines}. 

\subsection{From edge, fog \& cloud to AI}
Currently, there are three main approaches to providing AI-enabled applications: edge, fog, and cloud AI. Edge AI \cite{wang2020edge} involves deploying AI applications directly on the devices where data is generated. This approach allows users to utilize moderately advanced hardware solutions that are in close proximity to the data that requires processing. It is a cost-effective method that offers enhanced privacy. However, it lacks scalability when there is a growing demand for AI solutions.
On the other hand, cloud AI enables vertical scaling by leveraging increasing computing power as needed. Nevertheless, it faces limitations at the network level when dealing with substantial amounts of data. Fog AI offers a horizontal scaling solution by utilizing a centralized network of edge devices. It aims to fulfill the high computing requirements of cloud AI while reducing overall network traffic. Both fog and cloud AI adopt subscription-based business models, which may not be advantageous for businesses that do not require real-time AI solutions.
We propose a decentralized architecture that harnesses the computing capabilities of edge devices while maintaining low-latency connections between customers and \textbf{Naeural network nodes}. Our architecture provides an on-demand, micro-transaction payment-based option for clients with AI-powered end-to-end application needs.

\subsection{The Decentralized paradigm }

\subsubsection{The classic distributed computing}
Distributed computing has turned into a cost-effective, high-performance and fault-tolerant reality due to great technological advances and falling costs of hardware. It refers to a collection of independent entities (workers), organized in a clear hierarchy with worker nodes and a central authority overseeing them. These workers computationally cooperate to solve a problem that would otherwise be potentially intractable at individual worker level. 
A distributed system is characterized, according to various authors \cite{ajay2008distributed}, by the following features: (a) no common physical clock - which guarantees asynchronicity amongst the processors, (b) no shared memory - which means that message-passing is required for communication, (c) autonomy and heterogeneity - the processors are not part of a dedicated system but cooperate with one another to solve the problem jointly; they have different speeds and can be running a different operating system.

\subsubsection{Transition to blockchain in distributed computing}
Classic distributed computing systems are typically centralized and rely on a central authority to manage and process data. Whereas, Distributed Ledger Technology (DLT) \cite{burkhardt2018ledger} is a specific type of distributed computing that is characterized by its use of a decentralized and immutable ledger to record and share data across a network of nodes.

One of the key differences between distributed ledgers and classic distributed computing is the way data is stored and shared. In a distributed ledger, transactions are recorded on a public or private blockchain and then replicated across the network of nodes while in usual distributed computing scenarios data is centrally stored and processed. Thus, in decentralized setting. a tamper-proof record, which can be accessed by anyone on the network, of all transactions is possible while in the classic distributed case the centralized date is prone to unwanted modifications. Even more, in almost any  implementation, distributed ledgers often use consensus mechanisms, such as those based on \emph{BFT}\cite{castro1999practical}, to ensure that all nodes on the network agree on the current state of the ledger. This guarantees that the data stored on the ledger is accurate and secure. 

Therefore, while in distributed computing the system relies on a main or master processing entity in the decentralized setting a peer-to-peer organization replaces the server-workers architecture.

\emph{BitTorrent} and \emph{IPFS (InterPlanetary File System)}\cite{benet2014ipfs} are two examples of how traditional distributed computing systems are being integrated with blockchain technology to create decentralized, secure, and efficient solutions for data sharing and storage.

\emph{BitTorrent} \cite{bittorent}, one of the first decentralized protocols for file sharing, uses a peer-to-peer network to allow users to share files directly with each other, without the need for a central server. This greatly increases the speed and scalability of file sharing. In recent years, \emph{BitTorrent} has been integrated with blockchain technology to create a decentralized file storage and sharing platform called \emph{BitTorrent File System (BTFS)}\cite{btfs}. \emph{BTFS} allows users to share files and earn rewards in the form of cryptocurrency for contributing storage and bandwidth to the network.

Similarly, \emph{IPFS} \cite{benet2014ipfs}\cite{ipfs} is a peer-to-peer protocol for sharing files and data across the internet. It allows users to share files directly with each other, rather than rely on a central server. \emph{IPFS} has also been integrated with blockchain technology to create a decentralized and distributed content delivery network called \emph{Filecoin} \cite{filecoin}. \emph{Filecoin} allows users to rent out their unused storage space and bandwidth to other users in exchange for \emph{Filecoin} tokens.

\subsubsection{AI \& Blockchain}

AI and blockchain technology can complement each other in a number of ways. One major benefit of using blockchain in conjunction with AI is the ability to securely and transparently share and access data. All actions, some of which could be expensive modeling, human data creation/annotation or even a simple AI-app-generated event, are traceable on a public distributed ledger. This can be particularly useful in areas such as medical research, where large amounts of sensitive data need to be shared among multiple organizations. Another benefit is that blockchain can be used to create decentralized platforms for AI model training and deployment, as described in a framework proposed by \emph{Microsoft Research}\cite{DBLP:journals/corr/abs-1907-07247}. This can help ensure that the models are fair and unbiased, as well as provide a way for individuals and organizations to share and monetize their own models.

A few examples of companies that are combining blockchain with AI include \emph{Ocean Protocol}\cite{oceanprotocol}, that uses blockchain to create a decentralized marketplace for data, and \emph{SingularityNET}\cite{singularitynet}, which fosters blockchain to create a decentralized platform for AI services.

Below, we list a few industries/sectors and benefits brought about by AI and blockchain.

Supply Chain Management: The combination of AI and blockchain can be used to create a more efficient and effective supply chain management system. For example, deep learning algorithms can be used to analyze data on supply chain transactions and optimize processes, while blockchain can be used to create a secure and transparent record of every transaction.

Financial Services: The combination of AI and blockchain can be used to create a more secure and efficient financial system. For example, deep learning algorithms can be used to detect and prevent fraudulent activities, while blockchain can be used to create a tamper-proof record of every transaction.

Healthcare: The combination of AI and blockchain can be used to create a more personalized and secure healthcare system. For example, deep learning algorithms can be used to analyze patient data and provide personalized treatment plans, while blockchain can be used to create a secure and privacy-preserving record of every patient's medical history.

Energy Management: The combination of AI and blockchain can be used to create a more efficient and sustainable energy management system. For example, deep learning algorithms can be used to analyze data on energy production and consumption and optimize energy use, while blockchain can be used to create a transparent and auditable record of every energy transaction.

Decentralized Autonomous Organizations (DAOs): The combination of AI and blockchain can be used to create more effective and autonomous DAOs. For example, deep learning algorithms can be used to analyze data on user behavior and preferences and optimize product recommendations and investment decisions, while blockchain can be used to create a transparent and auditable record of every transaction within the organization.

\subsubsection{Integrating AI on blockchain through smart contracts and microtransactions}
Smart contracts and microtransactions are two key ways that blockchain technology can enable the integration of AI on the blockchain, by providing an infrastructure for automating transactions and for monetizing AI services.
Smart contracts are self-executing contracts with the terms of the agreement written directly into code. 
They can be used to automate the execution of transactions when certain conditions are met, without the need for intermediaries.

This is particularly useful in the context of AI, as smart contracts can be triggered by AI systems to execute transactions automatically. 
For instance, an AI model may make predictions about the price of a stock. If the conditions set in the smart contract are met, the trade will be executed automatically.

Microtransactions, on the other hand, are small financial transactions that can be executed efficiently and inexpensively on the blockchain. They can be used to enable the monetization of AI services on the blockchain. For example, an AI model that is trained on a decentralized network could be made available to other users in exchange for small amounts of cryptocurrency. This would allow the creators of the model to earn income from their work, and it would also encourage the sharing and improvement of AI models.

\section{Architecture}

\subsection{A top-down view}

\begin{figure}[htp]
    \centering
    \includegraphics[width=16cm]{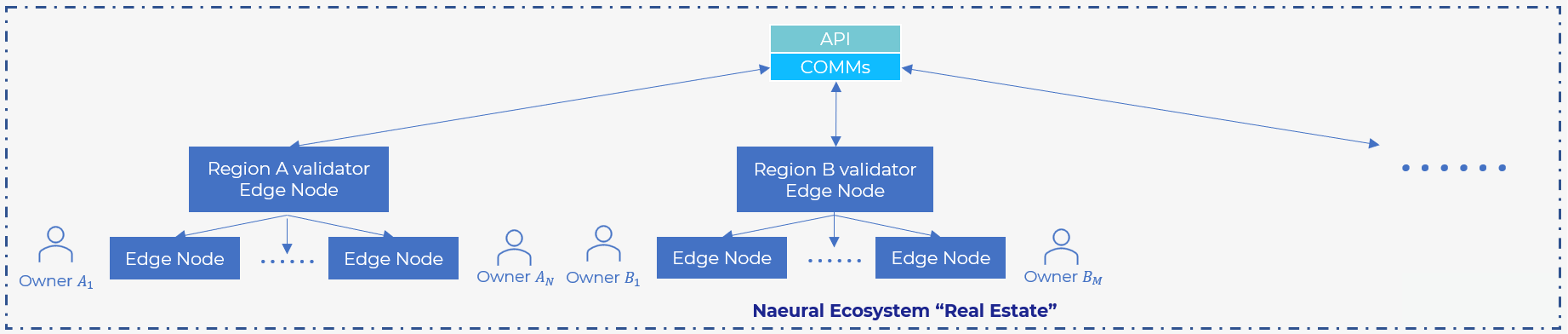}
    \caption{\textit{Overall view of the Naeural E2s network of networks}}
    \label{fig:aixp_overall}
\end{figure}

The proposed decentralized architecture, as illustrated in \textbf{\figurename{\ref{fig:aixp_overall}}}, is grounded in the well-established design of validator-like \textbf{Naeural Processing Nodes}, which facilitate regional or global aggregation of \textbf{Naeural Processing Node Pools}. Distinct from traditional validators responsible for blockchain protocol consensus, the primary function of an \textbf{Naeural Validator Node} pertains to job validation and communication aggregation. Concurrently, all \textbf{Naeural Processing Nodes} participate in a private blockchain consensus, ensuring secure and immutable job execution and data transfers.

\begin{figure}[h]
    \centering
    \includegraphics[width=16cm]{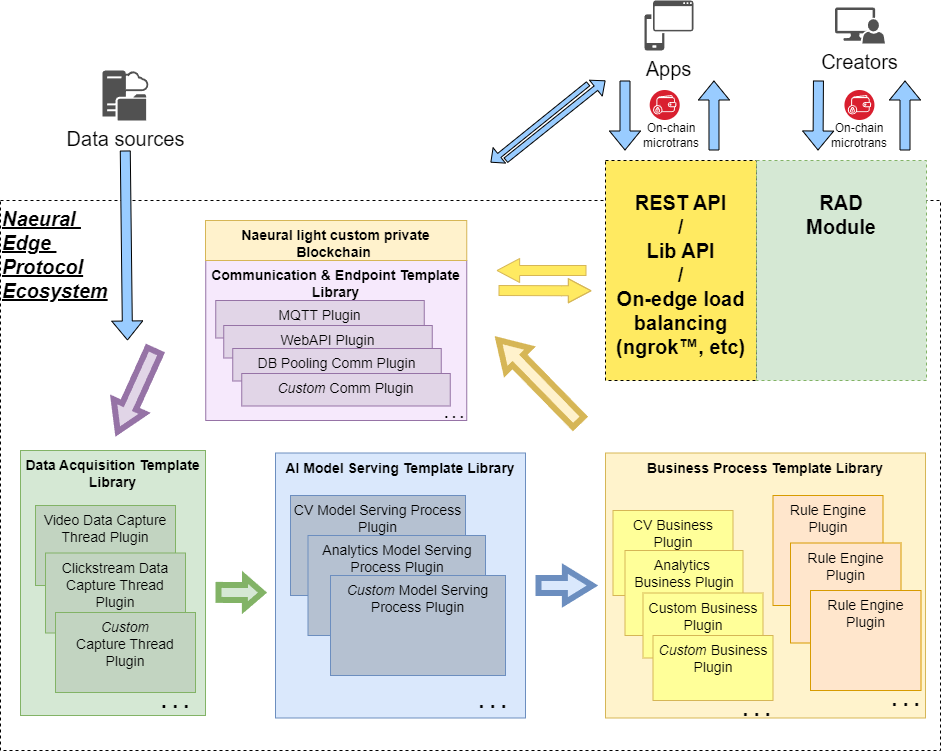}
    \caption{\textit{Top view of Naeural Execution Engine ecosystem}}
    \label{fig:ee}
\end{figure}

The overall \textbf{Naeural Ecosystem} encompasses various middleware components, including IoT message broker tools and REST APIs for diverse use-cases. At its core is the \textbf{Naeural Execution Engine (E2)}, depicted in Figure \ref{fig:ee}, along with its platform-agnostic SDKs. The \textbf{E2} 
can be deployed on a wide array of target operating systems and hardware devices, with or without advanced GPU compute capabilities. \textbf{Naeural} employs both a private proprietary lite blockchain for internal use and an external public mainnet blockchain for microtransactions orchestrated by individual \textbf{E2} processing nodes. Consequently, each \textbf{E2} is uniquely identified and recognized within the network through a public blockchain (mainnet) non-fungible utility token. The subsequent paragraphs will discuss the primary decentralized processing distribution and the integration of a microtransaction-based worker reward system.

Although a comprehensive tokenomics strategy is proposed, it has not yet been publicly released. This paper focuses on the high-level technological aspects related to the \textbf{Naeural E2 Node Deeds}' interaction with the protocol, the fundamental principles of oracle-like communication between the \textbf{mainnet blockchain (BC)} and \textbf{Naeural E2s}, and the overall algorithmic approach for reward distribution \ref{alg:joballoc}\ref{alg:jobexec}\ref{alg:rewards}.

In \textbf{\figurename{\ref{fig:aixp_overall}}} the regional validation nodes coordinate the communication between nearby local nodes as well as provide inter-region distributed ledger data synchronization.

The \emph{Proof-of-AI protocol (PoAI)} \ref{alg:rewards} manages the rewarding of users who operate \textbf{Naeural E2} processing nodes. This incentivizes the provision of compute power by both miners and service consumers. Miners use the \textbf{Naeural E2} solely for decentralized job execution, while service consumers use specific functionalities or applications developed by \textbf{Naeural} content creators and service providers. As previously mentioned, an in-depth explanation of public on-chain algorithms and procedures is beyond the purpose of this paper, however the pseudo-code algorithmic approach of \textbf{Naeural Proof of AI (PoAI)} can be analyzed in the current paper.

In order to summarize the funds allocation and distribution in our job execution agnostic reward pools we will denote: $T_{N}$ the total time a given \textit{node} $N$, identified by a unique smart-contract based on a \textit{Node Deed} in the overall population of all $A$ \textbf{Naeural Processing Nodes}, is alive in the network; $T_{epoch}$ the protocol epoch time which is the interval used to calculate and distribute a round of rewards to all active nodes; $E$ the total number of epochs since the protocol genesys; $P_{N_E}$ the power score of a given node $N$ during epoch $E$. The final formula of the reward $R_{N_E}$ of the $Node$ $N$ at epoch $E$ is presented below in \equationautorefname{\ref{eqr:5}}. 

\begin{align}
T_P &= T_{epoch} * E\label{eqr:1}\\
\tau_{N} &= \frac {T_{N}}{T_P}\label{eqr:2}\\
I_{N_E} &= e^{P_{N_E}} * \tau_{N_{E}}\label{eqr:3}\\
I_{E} &=\sum_{X \in A}{\frac {\tau_{X} * e^{P_{X_E}}}{T_P}}\label{eqr:4}
\end{align}

An important observation is that the power score $P_i$ for any node $i$ - see equations \ref{eqr:3} and \ref{eqr:4} - can have both negative as well as positive values - i.e. the power score can get penalized in the negative range.

\begin{align}
R_{N_E} &= \frac{I_{N_E}}{I_{E}} \label{eqr:5}
\end{align}

The generic pseudo-code that depicts the proposed allocation of token rewards based on the previous equations \ref{eqr:1},\ref{eqr:2},\ref{eqr:3},\ref{eqr:4} and \ref{eqr:5} can be found in \textit{Algorithm \ref{alg:joballoc}}. Worth mentioning is that, although ideally the proposed architecture relies on real online on-chain micro-transactions that incentivize the participants in a entirely trust-less fashion, the system can work without any problems in isolated environments. In such private, educational sandboxes or particularly isolated networks the reward tokens are treated as proofs of historical participation in the decentralized distributed processing.

\subsection{End-to-end low-code pipelines}
Developing end-to-end applications is a time-consuming process due to its multiple high-complexity stages. These stages include tasks such as developing and deploying communication modules and endpoints, implementing data acquisition streams, and developing advanced data processing techniques using AI-based algorithms or complex heuristics. Additionally, it involves establishing a continuous deployment strategy and executing it effectively. Usually multiple teams handle all these different software development streams using different skills and even different education backgrounds. 

The \textbf{Naeural Ecosystem} aims to simplify the process of developing and deploying complex application systems by dividing it into four distinct heterogeneous subsystems: communication, data acquisition, model serving, and business process as seen in \figurename{\ref{fig:ee}. Each subsystem includes its own highly composable plugins, each with configurable parameters that allow customization of functionality without requiring additional code to be written. This versatility is achieved by allowing a pipeline to be configured by specifying which plugins to use and how they should be configured, enabling easy alteration of the pipeline's functionality without code intervention or with an extremely low code footprint. This structure allows developers to use pre-existing or create custom low-code solutions for each of the subsystem within the network's ecosystem, simplifying development, deployment and integration of complex application systems.

\subsection{Secure execution with Plugin API}

Due to the nature of trustless decentralized distributed job execution, multiple security issues arise: (a) custom code safety, (b) request and result message immutability, (c) data confidentiality and (d) solution proofing.
Challenges (b), (c) and (d) are cryptographic issues that can be addressed either with the proposed distributed ledger architecture or with neural model training and inference based on homomorphic encryption. The former relies on code safety checking and signing.

Two primary types of business plugins have been developed, namely secure plugins and user plugins. These plugins are currently built using \emph{Python}\cite{vanrossum1995python}, a high-level general-purpose programming language. Business plugins can be considered either generic templates or default plugins for preparing business insights data/reports. On the other hand, secure plugins encompass a range of user plugins designed as proposals for future secure services, as well as code snippets created by anyone using the \textbf{Naeural E2} within the trustless network.

Both deployed user plugins and custom user code snippets are regarded as potentially unsafe due to their nature. Therefore, they are subject to limitations in terms of programming language keywords, functions, and packages. They must adhere to a specific higher-level Plugin API and undergo hash-sign verification, profiling, and code re-checking to prevent code injection and other tampering processes each time distributed execution is requested. This safety measure is implemented regardless of whether the code is transferred and executed either peer-to-peer or remotely by the same initiator and worker.

It is important to note that all these approaches can be accessed and utilized through external endpoints. This can be done either via a \emph{Python} client that utilizes our \textbf{Naeural E2 PyE2} package or other similar approaches regardless of the programming language.

\subsection{Multi-framework serving processes}

The \textbf{Naeural Network} and its ecosystem are designed to support applications powered AI. This is achieved by incorporating various AI models through the use of AI serving plugins within the serving subsystem. Additionally, the ecosystem is designed to accommodate applications that do not rely on AI models or techniques but are well-suited for decentralized environments.

To enable this, \textbf{Naeural} provides a serving plugin interface, illustrated in Figure \ref{fig:sc_uml}, which is not limited to specific frameworks like \emph{TensorFlow} \cite{abadi2016tensorflow}. In this setup, all data input, inference, and prediction output are handled within the main thread of the main process. The implementation of the serving plugin interface primarily involves model loading, data pre-processing, and inference post-processing.

Consequently, the serving layer of the execution engine consists of multiple individual processes, with each process dedicated to running inference for a single model, whether it is a single-stage or multi-stage model. This allows for parallel execution of all the model-serving processes, ensuring that each process is encapsulated, easily monitored, and managed without interference.

\begin{figure}[h]
    \centering
    \includegraphics[width=12cm]{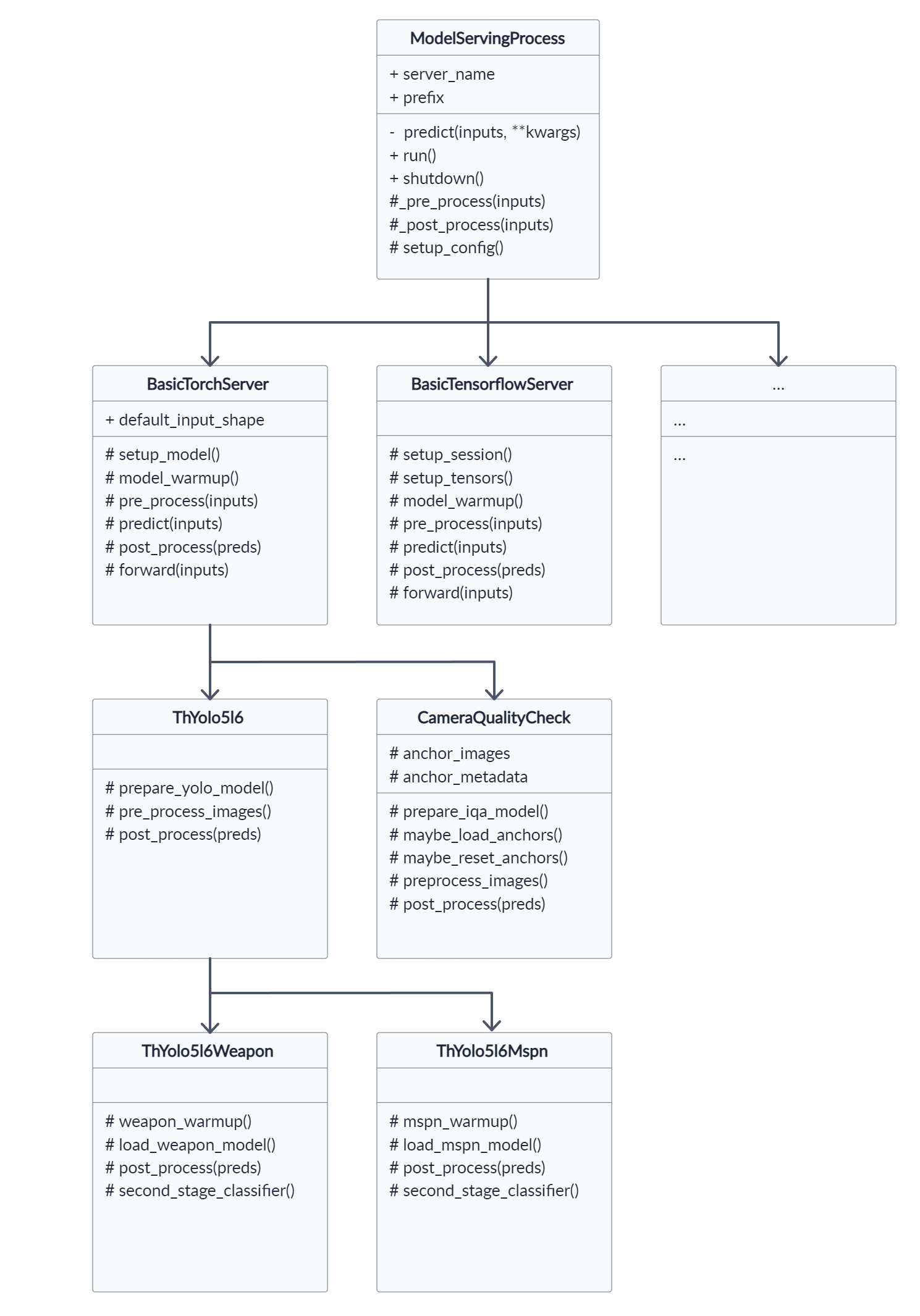}
    \caption{\textit{High-level Naeural E2 model serving processes UML diagram}}
    \label{fig:sc_uml}
\end{figure}

\subsection{Distributed decentralized execution}

The architecture of the \textbf{Naeural Network} is specifically designed to achieve scalable and decentralized distribution of different processing workloads. This is accomplished through a combination of peer-to-peer trustless job execution based on \emph{Map Reduce} \cite{dean2008mapreduce} and \emph{Scatter Gather} \cite{cutting2017scatter} techniques, along with DLT. Throughout the design process, ensuring the security of all participants has been of paramount importance, influencing the network architecture and the distribution of processing workloads.

The ultimate objective is to empower a diverse range of participants, including service providers, work offload providers, and consumers, enabling them to fully leverage the capabilities offered by the network.

To provide a concise overview of the high-level process, we define an instance of the \textbf{Naeural E2} that requires external processing offload as a \textbf{Main Node}, while other nearby \textbf{Naeural E2} instances will be referred to as potential \textbf{Worker Nodes}. In this context, the \textbf{Main Nodes} have the capability to initiate distributed execution pipelines, each comprising two main processing stages, each with two distinct steps.

The initial stage, known as the \textbf{Map Stage}, is responsible for (i) searching for suitable nodes among all available \textbf{Workers} and (ii) mapping and assigning jobs to individual nodes. Following the \textbf{Map Stage}, the subsequent stage, referred to as the \textbf{Reduce Stage}, involves (iii) overseeing the completion of worker jobs and ultimately (iv) compiling and processing the results.

In order to achieve scalability within the trustless environment, it is crucial to address the security concerns associated with each stage (\textbf{Map \& Reduce}) while maintaining coherence with the functional aspects of the distribution pipeline. To accomplish this, certain measures are implemented.

During the \textbf{Map Stage}, nodes are required to publish both their system specifications and job assignments to the blockchain. This ensures the immutability of the data. A \textbf{Worker Node} must sign and commit to its capabilities, as these factors impact its \textbf{Power Score} and, consequently, its share of profits.

In the \textbf{Reduce Stage}, \textbf{Workers} periodically send their job statuses along with \textbf{Zero-Knowledge Proofs} \cite{goldwasser2019knowledge}. These proofs allow the \textbf{Main Node} to verify, with minimal computational effort, that overall progress is being made over time, without revealing sensitive information. This process ensures that the \textbf{Main Node} can assess the increasing progress of the tasks without compromising data privacy and security.

The primary goal of the \textbf{Naeural Ecosystem} is to cater to a broad range of participants, including content creators, processing node operators, software developers, and end-consumers. To achieve this objective, \textbf{Naeural} provides a comprehensive platform that is adaptable, reliable, secure, and easily scalable. This platform is designed to support not only AI-powered applications but also other decentralized applications, regardless of whether they utilize AI models or not.

Through its compatibility with various frameworks and the ability to execute multiple model-serving processes in parallel, \textbf{Naeural} ensures an efficient and flexible solution for its users. The platform's distinctive design, coupled with its integration with blockchain technology, enables secure and transparent operations, effectively meeting the diverse needs of participants.

To enhance the adoption of the ecosystem by the public, software development kits (SDKs) have been developed. These SDKs support multiple programming languages and frameworks, such as \emph{Python}, significantly expanding integration options and improving the user experience. By using the SDKs, users gain access to the \textbf{Naeural Network} through a partially functional paradigm, which offers callbacks for manipulating payloads received from \textbf{Naeural Nodes} and provides simple yet powerful features for communication with the network. This empowers users to connect to the network, establish new end-to-end pipelines or join existing ones, create predefined or custom jobs within the pipeline, and receive notifications and heartbeats, facilitating seamless interaction with the ecosystem.

The \textbf{Naeural SDKs} are intended to be used as a cohesive integration tool between the AI processing network and any existing or new software applications. They are designed to automate and manage the lifecycle of requests made on the \textbf{Naeural Network}. \textbf{Naeural SDKs} come in several flavors targeting either the \emph{Python} or other developers' communities.

In the annex of this paper, we present a simple Python 3.x example \ref{lst:pyee} that demonstrates the process of connecting to a processing \textbf{Naeural E2 Node}, executing tasks remotely, and distributing an end-to-end pipeline. However, it's important to note that this example does not include the necessary steps to enable Proof-of-AI (PoAI) reward distribution, which involves compensating the processing node pool for the executed job.

\subsubsection{The Naeural SDK}

To support developers in integrating network functionality into their solutions, a Command Line Interface (CLI) tool is provided - namely \textbf{nepctl}. This tool assists in various tasks such as obtaining available online nodes within \textbf{Naeural Network} or configure the SDK installation. 

\begin{quote}
Note that each individual SDK will have a private key generated upon the first installation on the target computer. This private key allows the SDK client to connect to the Naeural network and even pair the local SDK with certain nodes by explicitly adding the SDK public key in the edge node allowed list.    
\end{quote}

The application runtime incorporates a package tailored for a specific programming language, which is meant to be seamlessly integrated into the business process services of the consuming application. The implementation details may vary depending on the programming language - currently supporting Python - but generally involve callbacks which trigger internal events. The responses obtained from these events are then analyzed and converted into data structures that are specific to the programming language in use. This allows developers to design and create business objects that align with the requirements and objectives of the application.

For Python development teams, the \textbf{Naeural PyE2 SDK} package can be installed and once this prerequisite is fulfilled, the imported libraries within the project will handle all validation and transformations between the application's business objects and the expected message formats of the \textbf{Naeural Nodes}.

To publish a message on the network, it is essential to establish a valid communication channel with a public \textbf{Naeural} \emph{MQTT}. This can be achieved by connecting to the \textbf{Naeural Validator Node} or by using the local broker bundled with the \textbf{Naeural Node}. Both options provide extensive availability of resources and published AI services. Furthermore, \textbf{Naeural PyE2 SDK} provides built-in support for listening to payloads, intercepting errors, and diagnosing any issues that may arise due to incorrect use of the AI pipeline plugins.

\subsection{Integration with Web3}
The use of on-demand AI functionalities and capabilities offered by the \textbf{Naeural Network} involves two main stages. The first stage involves the financial commitment of the processing power consumer, where a designated amount of tokens is allocated for covering AI request-related expenses. In the second stage, decentralized job execution takes place.

To facilitate efficient communication between local nodes, the \textbf{Naeural Network} employs the \textbf{Naeural} communication protocol, which utilizes \emph{MQTT} as the underlying transport layer. \emph{MQTT} is a widely adopted protocol known for its reliable scalability in IoT applications. The lifecycle of the \textbf{Naeural Protocol} is encapsulated within the \textbf{Naeural PyE2 SDK}, simplifying its implementation and usage.

To guarantee compensation for potential worker nodes' efforts, it is essential that network requests are accompanied by a financial commitment. By including information pertaining to a valid inbound \textbf{Naeural} wallet transaction, an \textbf{Naeural Node} can process a designated workload. This step can be executed by the consumer application or via a separate procedure, as long as it is completed before submitting the workload to the network.

The \textbf{Naeural PyE2 SDK}'s primary roles are to manage relevant financial information before disseminating the request on the network and to simplify the integration of this technology within external systems. For processes that necessitate multiple network requests, substantial transaction fees may be incurred. To minimize expenses related to utilizing cryptocurrency networks, depositing a larger sum into an escrow account is possible, allowing the \textbf{Naeural} internal blockchain to oversee expended funds.

The \textbf{Naeural PyE2 SDK} employs specific \emph{Web3} components for interacting with payment channels, contingent on implementation. On the \textbf{Naeural E2} side, a simple validation is performed to determine whether the request is signed by the \textbf{Naeural Node} owner and run locally (in which case payment is unnecessary) or whether a valid funding transaction is attached.

The request entails disseminating a simple message to a user-controlled node or the nearest validator node. Subsequently, the network assumes responsibility for managing the request. Built with robust components, the \textbf{Naeural PyE2 SDK} offers a reliable communication channel with the AI-enabled node network. Request progression is monitored by event listeners within the SDK, and any outcome is documented and relayed to the initiating processes.

Throughout pipeline execution, \textbf{Naeural Network Nodes} generate status payloads, which outline the request's current state, report potential errors, and supply information on generated assets. These payloads are monitored by event listeners within the \textbf{Naeural PyE2 SDK} components, and appropriate handlers are provided for seamless interaction with ongoing processes.

\subsection{Naeural Rapid Application Development}

The \textbf{Naeural Ecosystem} envisions a future where AI-powered application development is accessible to individuals, regardless of their familiarity with programming languages. To achieve this, we are creating tools that enable both developers and non-developers to effortlessly articulate their business requirements and translate them into instructions compatible with the \textbf{Naeural SDK}.

For users with existing programming experience, our initial set of tools is designed to facilitate expressing their business needs. We provide code samples that demonstrate the functionality of the \textbf{Naeural Ecosystem}. To streamline the process, we offer a code-sample gallery tool, allowing users to swiftly run samples, modify code and inputs, and test various results without needing to launch an IDE or create a new project.

To accommodate individuals without programming expertise, we are developing low-code tools that empower these users to generate their own code samples without extensive programming experience. Our ultimate goal is to provide a user-friendly interface that enables anyone, irrespective of their familiarity with source code and programming, to use drag-and-drop tools on-screen to express their desired business functionality.

AI-powered applications rely on data, and incorporating data from existing sources into applications is a crucial part of development. This typically requires previous programming experience. We aim at eliminating this requirement by providing visual connectors, allowing users to easily integrate existing data into their applications through a drag-and-drop graphical interface.

While multiple approaches already exist on the market, such as \emph{Azure ML} \cite{barnes2015azure} and \emph{Amazon SageMaker} \cite{liberty2020elastic}, these tools necessitate extensive knowledge of Data Engineering and Data Science. Our aim is to provide simpler visual means that will be usable in the decentralized ecosystem.

Regardless of one's expertise, we are convinced that visual tools have the potential to greatly enhance the developer experience.

\section{Applications}

In this section of the paper, we present a range of applications that have been effectively implemented within the \textbf{Naeural Network}. These applications cover diverse domains and industries, catering to both our internal usage and our customers' requirements. Additionally, we have collaborated with external partners to develop applications that are specifically tailored for their needs.

While most of these use cases are presently deployed and operational across multiple private or public worker pools that utilize \textbf{Naeural E2} on GPU systems, the distribution of workload does not currently incorporate PoAI-based incentivization.

\subsection{Physical security}
The average human attention span is approximately 10 seconds\cite{attention_span}, which decreases when an individual is tasked with monitoring multiple video streams simultaneously. In order to alleviate this strain, machine vision using deep learning models can be deployed in video surveillance to assist the operator in detecting unusual events. The multiple applications of this technology include industrial personal safety, location security, and video equipment security.

    \subsubsection{Industrial personal safety}
    This category of use-cases involves analyzing people's posture and body positioning using a multi-stage machine learning model and potentially applying specific heuristics. By using a deep learning model to localize persons in an image and then a second model to determine the positioning of specific body parts, we can gather valuable data about the actors in the scene. Furthermore, using second stage deep classifiers missing or improper protection equipment wearing can be alerted. Finally, complex correlation can be made using multiple AI models - i.e. such as determining specific postures or dangerous situations in industrial areas where, for example, heavy machinery is operated.
    
        \paragraph{Use-case: Fallen person}
        A problem that can occur because of the aforementioned attention issue, especially when the operator has to observe a bigger number of cameras, is the lack of immediate action taken when a person falls on the ground from various reasons and needs assistance. In some cases, a delay of even 5 minutes can make a substantial difference. As described above, we can use multi-stage models in order to ensure the safety of the working personal in a given environment. By analyzing the placing of some human joints and angles determined by those we can detect in a matter of seconds if an employee is lying on the ground and may need assistance.

    \subsubsection{Location security}
    This category involves analyzing the appearance of unusual elements or individuals in certain scenes, a use case specific for CCTV video analysis. To achieve this, we can use a deep vision detector that can be either class oriented - i.e. for detecting a animal in a place where no animals are supposed to be - or even class agnostic for detecting suspect objects.

        \paragraph{Use-case: Vandalism}
        In this case a deep vision model is trained to detect either subtle or significant changes in a scene and thus can be used to identify signs of vandalism, such as graffiti on a wall, trash on the floor, moved furniture or even unknown objects dropped in the target scene.
        The integration of AI in video surveillance systems not only improves the efficiency of detecting vandalism, but also enables prompt intervention and mitigates potential security risks.

    \subsubsection{Video equipment security}
    While the previous category analyzes the scene in a given image, this third category analyzes the image itself in order to detect potential damage suffered by the recording equipment, either by natural causes (e.g. outdated cameras) or by human intervention (e.g. moved cameras). By analyzing the positioning of certain edges in the frame or other features provided we are able to quickly notify the technical staff in order to solve the issue at hand.

        \paragraph{Use-case: Camera quality check}
        In this case a deep vision model is employed to identify minor deviations in the placement of edges or abrupt fluctuations in frame entropy, thereby detecting any potential damage or malfunction in the video cameras. By leveraging these features, the model is able to promptly alert technical personnel to any issues that require immediate attention. This use-case not only guarantees the dependability and efficiency of the surveillance system, but also saves valuable time and resources that would otherwise be spent on manual monitoring and equipment inspection.

\subsection{Predictive analytics}
Predictive analytics is the use of data, statistical algorithms and machine learning techniques to identify the likelihood of future outcomes based on historical data \cite{klimberg2016fundamentals}. Such techniques are used by a large variety of businesses - from retail brick and mortar stores to physical security companies in order to optimize their business processes or to make predictions regarding the future behaviour of people or equipment.

\subsubsection{Replenishment}
One of the biggest problems of retail logistics is that of over-supply and under-supply. Over-supply can incur losses based on the perishability of the over-supplied products, as the stock is not sold in time or based on the fact that storage is also a cost; while under-supply may induce missed potential sales due to the product not being in stock.

\paragraph{Use-case: Predictive replenishment}  The easiest, but not the most efficient solution, is the use of heuristics \cite{stefanovic2015collaborative} to generate replenishment orders. The main drawback is that good results require very complex algorithms which are hard to maintain and do not easily scale. The alternative to the heuristic method is the automated learning approach. Machine learning techniques, especially deep representation models \cite{kilimci2019improved}, are able to more accurately find patterns in historical data and using continuous training methods can easily adapt to new circumstances in order to provide reliable consumption predictions.

\subsubsection{Stock prediction}
Traditionally, the user interaction with a product is an integral part of the product itself. Therefore, significant effort is dedicated to ensuring a streamlined user journey where all available choices are visible and easily discernible. This involves employing established user experience patterns for menus, buttons, layouts, and other components that facilitate the manipulation of data within a product.
AI presents a distinctive challenge. The AI product itself cannot be directly experienced, as users can only interact with the resulting output — an amalgamation of vast amounts of data. This characteristic allows for seamless integration of AI into other products. 
\paragraph{Use-case: Stock prediction in a spreadsheet}
An intriguing application of this fusion between AI and everyday operations is the inclusion of prediction functionality within a spreadsheet. To illustrate this application, we can demonstrate how the capabilities of the \textbf{Naeural Network} can be leveraged. A user selects a specific range of cells within the spreadsheet that contains relevant values for the desired prediction. These values are then transmitted to an API that incorporates \textbf{Naeural Nodes}. The API performs validation on the received data, formats it for communication with the nearest \textbf{Naeural Node}, and awaits the response generated by the predictive model. Finally, the response is relayed back to the spreadsheet for display.

The web-API that serves the predictions in this example relies on the \textbf{Naeural Node} internal FastAPI-based plugins that support both RESTful architecture as well as statefulness.

\subsubsection{Use-case: Predictive Maintenance}

A prominent use-case for deploying predictive analytics applications within decentralized infrastructures is industrial predictive maintenance. Predictive maintenance refers to the proactive identification and resolution of potential failure points in industrial equipment before critical malfunctions transpire \cite{mobley2002introduction}. This is typically achieved by monitoring equipment behavior changes and comparing them to the historical behavior of similar items.

Traditional solutions for predicting equipment breakdown, such as autoregressive models like \emph{ARIMA} \cite{kanawaday2017machine}, often suffer from low accuracy and an inability to detect subtle changes in equipment behavior. Consequently, deep learning-based methods \cite{zhao2019deep} are employed to create accurate and scalable systems. These deep models can effectively recognize patterns, predict future malfunctions, and readily generalize to new situations, offering a reliable solution.

\subsubsection{Use-case: Anomaly Detection}
Closely related to predictive maintenance, anomaly detection is another crucial use-case. Anomaly detection involves the identification of data points that significantly deviate from the majority of the data \cite{chandola2009anomaly}. It can be applied in physical security applications to detect potential security threats or system malfunctions. Such systems are essential due to the high volume of telemetry and events, which are costly to parse by human operators and prone to errors due to the tedious nature of the task. A common solution to the anomaly detection problem is the use of \emph{Multivariate Gaussian Models} to identify suspicious events \cite{mehrotra2017anomaly}. While this method has low computational cost and is easy to implement, it suffers from low accuracy and an inability to represent and leverage complex data patterns. Alternatively, deep representation models \cite{chalapathy2019deep} can be employed to take advantage of higher-order patterns, enhancing the accuracy and effectiveness of anomaly detection systems.

\subsection{Other applications}
\subsubsection{Decentralized game engine}
In the previous examples, the primary focus was on use-cases that employ artificial intelligence particularly shallow and deep machine learning models. Nevertheless, the ecosystem is also capable of supporting the development, integration and even migration of non-AI applications given the hypothesis that the particular target application is suitable for running in a decentralized environment. To demonstrate this versatility, a simple sandbox simulation life-game was designed showcasing how virtual agents interact with one another in an evolutionary heuristic environment. While the initial purpose of this sandbox game was to illustrate the flexibility of \textbf{Naeural Development Ecosystem}, this mini-project can be used as a tutorial or even as a basic application template. 

While being based on the architecture presented in Figure \ref{fig:ee}, most of the development effort has been geared towards three areas: (i) sandbox simulation encapsulated as a \textbf{Naeural E2} model serving module,(ii) the business logic that prepares the pre-rendering vector data and finally (iii) the UI that renders and enables interaction with the sandbox game. 
The model serving component implements the life-game simulation stateful process itself instead of a traditional stateless machine learning model. While in a conventional model serving process the inferred data is based on input data generated from an upstream data acquisitor, the simulation process here produces the state of the environment after each cycle. The business heuristics component is then responsible for preparing the data that will be delivered from \textbf{Naeural E2} to an external custom user interface. Furthermore, the custom UI consumes the inflow of data while generating interaction events in turn that are delivered to the \textbf{Naeural Processing Node} which runs the simulation game.

\section{Ongoing work}

Our development teams are currently working on several crucial areas aimed at addressing key aspects of decentralized execution. These areas include optimizing AI inference and training processes by leveraging homomorphic encryption through our \textbf{Naeural EDIL Framework} (EDIL - Encrypted Decentralized Inference and Learning) . Our framework currently supports popular deep learning frameworks such as \emph{PyTorch} and \emph{TensorFlow}. Additionally, we are actively exploring the utilization of zero-knowledge proofs for evaluating job completion and refining essential workflows involving two-blockchains. This includes establishing efficient interactions between on-chain systems and the \textbf{Naeural E2} environment through oracles-based mechanisms. These ongoing developments demonstrate our commitment to advancing decentralized execution capabilities and enhancing the overall functionality and security of the \textbf{Naeural Ecosystem}.

\subsection{Distributed Homomorphic Encryption}

In decentralized MLOps, ensuring secure and confidential distribution of jobs and data manipulation is essential for maintaining privacy. Traditional encryption techniques that rely on cipher keys require the disclosure of original data to authorized parties, which is not ideal in a distributed system aiming to provide complete anonymization. To overcome this challenge, our proposed framework, known as \textit{EDIL}, introduces distributed homomorphic encryption (\textit{DHE}). By leveraging \textit{DHE}, we enable encrypted data processing without the need for decryption, ensuring both data confidentiality and security.

One of the significant advantages of \textit{DHE} is its ability to allow multiple parties to perform computations on the same encrypted data without revealing any information to each other. This characteristic makes it suitable for a trust-less environment where sensitive data must be managed securely without any disclosure.

Considering the computational demands of MLOps jobs, it is crucial to minimize the overhead introduced by homomorphic encryption to ensure scalability. To address this concern, our proposed \textit{DHE} approach utilizes a \emph{Domain Auto-encoder}-based method, which reduces the amount of information needed while retaining essential data features. From the perspective of a \textbf{Naeural Worker}, the encoder functions as a one-way function without the decoder, preserving the confidentiality of the data. However, to maintain minimal overhead and achieve the desired secrecy, a unique auto-encoder is required for each job type. This approach allows for secure and efficient manipulation of sensitive data within a decentralized ML environment.

\subsubsection{Decentralized Federated Learning}
To dispatch a training job to \textbf{Naeural Worker Nodes}, the self-designated \textbf{Naeural Main Initiator Node} either automatically trains a \textit{DHE} domain auto-encoder with the data to be processed or uses a predefined one from an existing library within the \textbf{Naeural Ecosystem}. This generates an encoder compressing the data's salient features. The \textbf{Initiator Node} can then define the end-to-end training job, including the machine deep learning model that consumes the domain encoder output and sends the compressed and DHE data to each worker-peer. Subsequently, workers utilize this \emph{DHE} approach to train the model using the encoded information in multiple rounds, similar to \emph{FedAvg} \cite{konevcny2016federated}, under the supervision of the \textbf{Naeural Initiator Node}.

\subsubsection{Encrypted Inference}
For deploying an inference job to \textbf{Naeural Nodes}, the \textbf{Naeural Initiator Node} or external API first encodes the raw data using a neural model stub or a pre-trained neural encoder in compliance with the proposed \emph{DHE} approach. The \textbf{Initiator Node} then sends both the \textbf{Main} neural model and the encoded data to the partner nodes accepting the job. The receiving nodes forward the input data through the distributed \textbf{Main} model and execute specific post-processing heuristics and code required by the \textbf{Initiator Node} before sending the final result. Throughout this process, receiver nodes only have access to the encoded format received from the \textbf{Initiator Node} as proposed by \emph{DHE}. By employing this approach, computational load can be distributed across a peer-to-peer decentralized ecosystem.

\subsection{Two layers of BC}
The \textbf{Naeural Network} utilizes a dual blockchain system, consisting of a public and private blockchain, to derive optimal benefits from micro-transactions in the financial domain as well as implement distributed data transactions immutability, respectively. It is imperative to note that while a single blockchain can potentially achieve these objectives, this approach is not favored as it would result in an undesirable gas cost associated with the commitment protocol for data-only transactions that should accommodate an arbitrary load with no correlated financial burden. 

The private blockchain is designed as a proprietary, lightweight system to guarantee operational freedom and minimize its architectural footprint. This blockchain can be employed in isolated, private settings and serves the primary function of enabling synchronization among the participating nodes in the network. This synchronization is crucial for the execution of delegated tasks and the signing of system specifications, both of which are integral to the operation of the \textbf{Naeural} distributed jobs. These processes enable trust-less operations in a heterogeneous operating environment as we aim to achieve a fully trust-less ecosystem for job execution and rewarding.

In contrast, the public blockchain relies on the existence of the private blockchain and is unable to operate independently. The primary objective of the public blockchain is to provide a straightforward yet powerful interface to facilitate the monetization of the entire system through a decentralized approach. This is achieved through the integration of an oracle between the public and private blockchain, which connects payments to job requests and the progress of those jobs. To further safeguard the integrity of the system, the public blockchain incorporates a democratic mechanism to deter fraudulent activities and promote transparency. For instance, participants within the public blockchain can issue challenges to one another in regards to job quality, with the outcome of these challenges resolved through a random selection of nodes, and recorded immutably on the blockchain for future reference. The challenge mechanism incurs costs supported by the challenger until their accusations are proven true, at which point they recover their funds.

\subsection{Oracles between on-chain and Naeural E2 worlds}

To attain the required level of automation within the ecosystem for multiple aspects from micro-transactions to node reliability history, the deployment of multiple oracles is necessary for the facilitation of communication between the on-chain and \textbf{Naeural E2}. The oracles enable off-chain to on-chain data transfer between the private and public blockchains in order to enable the allocation of jobs to\textbf{Naeural Worker Nodes} and the creation of worker incentivisation escrow pools. Furthermore, they enable sending commands to worker nodes and retrieve results, assist in computing proofs of job execution and enable on-chain publishing of the outcomes. As central components of the automated end-to-end process, oracles must be secured to prevent the occurrence of vulnerabilities that may have severe implications for the system. It is therefore of utmost importance to ensure that oracles behave in a fair and correct manner, even in scenarios where attackers attempt to compromise the system.

\subsection{Zero Knowledge Proof}
In order to enable trust-less data consumption and jobs within the decentralized distributed job execution ecosystem, it is crucial to implement a minimal \emph{Zero Knowledge Proof} \cite{goldwasser2019knowledge} framework and capability in the system. The use of \emph{zero-knowledge proof} allows the verification of the progress made by the \textbf{Naeural Worker Nodes} without disclosing any sensitive information regarding the progress of the job, where one only needs to verify that it is increasing, as well as allowing the job initiator to verify the actual completeness of a given finalized job in a feasible time. This not only maintains the confidentiality of the processed data but also minimizes network traffic by reducing the payload exchanged between the \textbf{Worker} and \textbf{Main Nodes}. In this way, the proposed system aims to provide users with a secure and efficient network for the execution of various types of distributed jobs.

\subsection{Conclusion}
The vision of the \textbf{Naeural} revolves around democratizing AI-powered application development and deployment as a transformative ecosystem to address multifaceted challenges and facilitate innovative solutions. All of our tools and technologies streamline the process of converting business requirements into instructions, enabling individuals with diverse backgrounds to seamlessly undergo this transformation.

In a broader context, the \textbf{Naeural Ecosystem} embodies the remarkable potential of blockchain technology in establishing a more efficient, and inclusive ecosystem, extending beyond the realm of AI. Through its decentralized nature, the ecosystem strives to promote greater autonomy, resilience, and equitable participation, offering a promising paradigm for future advancements in various domains.

\textbf{Naeural} serves as a testament to the potential of converting computing devices to real assets, while being a pioneer of AI democratization.

\bibliographystyle{unsrt}  
\bibliography{references}  

\newpage

\begin{lstlisting}[language=Python, caption={\emph{{Simple Python example based on Naeural PyE2 SDK where a local node is used to distribute a job}}}, label={lst:pyee}]

from pyee import Session

def custom_notifs_handling(self, notification):
  # notification object contains various information including
  # source node, context, etc
  return

def custom_heartbeat_handling(self, notification):
  # heartbeat object contains various information regarding nodes
  # that are reacheable within current processing pool
  return
  

pipeline_done = False
def pipeline_on_data(pipeline, signature, instance, data):
  # here we process on-data updates from the node that does the main
  # processing of our job and at some point we decide we finished our work
  pipeline.P(f'Received from box {pipeline.e2id} pipeline: {pipeline.name}:{signature}:{instance}')
  # etc ... at some point `pipeline_done = False`

if __name__ == '__main__':

  # create a session and connect to the local E2 comm server
  sess = Session(
    host="localhost", 
    port=1883, 
    user="naeural", 
    pwd="demopass",
    on_notifications=custom_notifs_handling,
    on_heartbeat=custom_heartbeat_handling,
  )
  
  sess.connect()
  
  # create a new pipeline on the 'e2id' box with one plugin instance
  n_workers = 3 # we ask the target E2 node to get help from 3 other E2s
  pipeline = sess.create_pipeline(
    e2id='stefan-e2', 
    name='test_normal', 
    type='custom_dedist_job', 
    code_path='./worker_job.txt', 
    config=[{'param_worker'+str(i) : i} for i in range(n_workers)],
    n_workers=n_workers, 
    on_data=pipeline_on_data
  )
  
  while not pipeline_done:
    # here we can do other main process 
    sleep(0.001) # yield is recommended  
    
  # close the pipeline (this also closes all processes opened on this pipeline)
  pipeline.close()
  
  # close the session
  sess.close()
\end{lstlisting}

\newpage

\begin{algorithm}
\caption{\textit{\textbf{NePRewards} - generic \textbf{Naeural} reward allocation end-to-end protocol}}\label{alg:rewards}
\textbf{Data/Inputs:}\vspace{1mm}\\
\hspace*{5mm} $N$  \Comment{N is the maximum number of nodes allowed by the protocol}\vspace{1mm}\\
\hspace*{5mm} $BC$ \Comment{\textit{Layer 1} or \textit{Layer 2} online block-chain protocol that will enable micro-transactions}\\
\hspace*{5mm} $EscrowPool$  \Comment{a \textit{smart-contract} in \textit{BC} holding unfinished jobs funds}\vspace{1mm}\\
\hspace*{5mm} $RewardPool$  \Comment{a \textit{smart-contract} in \textit{BC} holding finished jobs funds}\vspace{1mm}\\
\hspace*{5mm} $NodeDeeds[N]$ \Comment{The list of \textit{non-fungible} node-deed \textit{smart contracts} living in \textit{BC}}\vspace{1mm}\\
\hspace*{5mm} $MaxEpochTime$  \Comment{Seconds in a Naeural epoch }\vspace{1mm}\\
\hspace*{5mm} $Initialize$  \Comment{Initializes all smart-contracts}\vspace{1mm}\\
\hspace*{5mm} $MaybeUpdate$  \Comment{Updates (async and parallel) node deeds smart contracts using \textit{BC}}\vspace{1mm}\\
\hspace*{5mm} $CurrentTime$  \Comment{Seconds from 2000 or similar function}\vspace{1mm}\\
\hspace*{5mm} $GetNewJob$  \Comment{Generic function that retrieves a job data from posted jobs list.}\\
\hspace*{5mm} 
\begin{algorithmic}[1]
    \State $ProtocolState \gets GENESYS$ \Comment{This is the initial protocol \textit{genesys} stage}\;\vspace{1mm}
    \State $NodeDeeds \gets$ \textbf{call} $Initialize(BC)$\; \Comment{All deeds are initialized, although not all are distributed initially}\vspace{1mm}
    \State $NePEpoch \gets 0$;\vspace{1mm}
    \State $EscrowPool \gets 0;$\vspace{1mm}
    \State $RewardPool \gets 0;$\vspace{1mm}
    \State $ProtocolState \gets PROCESSING$; \Comment{We start distributing jobs and rewards stage}\;\vspace{1mm}
    \While{$True$}\vspace{1mm}
        \State $NePEpoch \gets NePEpoch + 1$;\vspace{1mm}
        \State $NodeDeeds \gets$ \textbf{call} $MaybeUpdate(BC)$;\vspace{1mm} \Comment{Allocation of deeds happening in other algo}\vspace{1mm}
        \State $StartTime \gets CurrentTime()$;\vspace{1mm}
        \While{$(CurrentTime() - StartTime) < MaxEpochTime$}\vspace{1mm}
            \While{$NewJobData \gets GetNewJob(BC)$} \textbf{\textit{in parallel}}\vspace{1mm}
                \State $Job, JobReward, Sender \gets NewJobData$;\vspace{1mm}\Comment{Job info is already available in BC}\vspace{1mm}
                \State $EscrowPool \gets EscrowPool + JobReward$;\vspace{1mm}
                \State \textbf{call async \textit{NePExecCheck}}\textit{(Job, JobReward, Sender)}; \ref{alg:jobexec}\vspace{1mm}
            \EndWhile\vspace{1mm}
        \EndWhile\vspace{1mm}
        \State $ActiveNodes \gets$ \textbf{call} $NePGetCurrentEpochNodeStatus()$;\vspace{1mm}
        \State $currNode \gets 0$;\vspace{1mm}
        \While{$RewardsPool > 0$}\; \Comment{\textit{RewardsPool} will be fully distributed among all participants }\vspace{1mm}
            \State $EpochActiveTime, EpochActivePower \gets ActiveNodes[currNode]$;\vspace{1mm}
            \State $currRewardPrc \gets \textbf{\textit{NePAlloc}}(currNode)$\ref{alg:joballoc};\vspace{1mm}
            \State $currReward \gets RewardsPool * currRewardPrc$;\vspace{1mm}
            \State $RewardsPool \gets RewardsPool - currReward$;\vspace{1mm}
            \State $NodeDeeds[currNode].addFunds(currReward)$;\vspace{1mm}
            \State $currNode \gets currNode + 1$;\vspace{1mm}
        \EndWhile
    \EndWhile\vspace{1mm}
\end{algorithmic}
\end{algorithm}

\begin{algorithm}
\caption{\textit{\textbf{NePExecCheck} - generic \textbf{Naeural} execution \& escrow release}}\label{alg:jobexec}
\textbf{Inputs:}\\
\hspace*{5mm} $Job$  \Comment{the \textit{job} request that will be distributed in the decentralized network}\vspace{1mm}\\
\hspace*{5mm} $JobReward$  \Comment{the proposed job fee payed by the \textit{Sender} of the job}\vspace{1mm}\\
\hspace*{5mm}  \Comment{Assumption is that the \textit{JobReward} has been already allocated in the \textit{EscrowPool}}\vspace{1mm}\\
\hspace*{5mm} $Sender$  \Comment{the \textit{owner} of the job request}\vspace{1mm}\\
\textbf{Data:}\\
\hspace*{5mm} $EscrowPool$  \Comment{a \textit{smart-contract} in \textit{BC} holding unfinished jobs funds}\vspace{1mm}\\
\hspace*{5mm} $RewardPool$  \Comment{a \textit{smart-contract} in \textit{BC} holding finished jobs funds}\vspace{1mm}\\
\hspace*{5mm} $JobStatus$  \Comment{a on-chain \textit{smart-contract} based matrix of jobs per \textit{Node}}\vspace{1mm}\\
\hspace*{5mm} $JobCount$  \Comment{a on-chain \textit{smart-contract} based vector of ongoing jobs per Node}\vspace{1mm}\\
\begin{algorithmic}[1]
    \State $JobCount[Sender] \gets JobCount[Sender] + 1$;\vspace{1mm}
    \State $JobID \gets JobCount[Sender]$;\vspace{1mm}
    \State $JobStatus[Sender][JobID] \gets IN\_PROGRESS$;\vspace{1mm} \Comment{JobStatus is stored in a on-chain smart-contract}
    \While{$JobStatus[Sender][JobID] == IN\_PROGRESS$}\vspace{1mm}
        \If {$Job.Status != NULL$}\vspace{1mm}
        \State $JobStatus[Sender][JobID] \gets Job.Status$;\vspace{1mm}
        \EndIf\vspace{1mm}
    \EndWhile\vspace{1mm}
    \If {$Job.Status == DONE$}\vspace{1mm}
        \State $EscrowPool \gets EscrowPool - JobReward$;\vspace{1mm}
        \State $RewardsPool \gets RewardsPool + JobReward$;\vspace{1mm}        
    \Else\Comment{\textit{Job.Status} is not ok due to sender cancellation}\vspace{1mm}
        \State $EscrowPool \gets EscrowPool - JobReward$;\vspace{1mm}\Comment {Funds extracted from escrow however...}
        \State \textbf{call} $LockFundsForReview(Job, JobReward, TIME\_24H);$ \Comment {the funds are locked pending \textit{Job} review}\vspace{1mm}
    \EndIf\vspace{1mm}
    \State $JobStatus[Sender][JobID] \gets NULL$;\Comment{Pop/Reset onchain storage data and free space}\vspace{1mm}
\end{algorithmic}
\end{algorithm}

\begin{algorithm}
\caption{\textit{\textbf{NePAlloc} - generic \textbf{Naeural} Node pool-share allocation}}\label{alg:joballoc}
\textbf{Inputs:}\\
\hspace*{5mm} $currNodeDeed$  \Comment{\textit{smart-contract} based unique identifier of the Naeural processing Node to be rewarded}\vspace{1mm}\\
\textbf{Data:}\\
\hspace*{5mm} $MaxEpochTime$  \Comment{Seconds in a Naeural epoch }\vspace{1mm}\\
\hspace*{5mm} $A$  \Comment{list of all Active Nodes based on oracle stored in a on-chain \textit{smart-contract}}\vspace{1mm}\\
\hspace*{5mm} $RewardPool$  \Comment{\textit{smart-contract} of the reward pool that acts as a escrow account}\vspace{1mm}\\
\begin{algorithmic}[1]
    \State $maxAlive \gets MaxEpochTime * NePEpoch$;\vspace{1mm}
    \State $AlivePrc \gets A[currNodeDeed].totalAliveTime$;\vspace{1mm}
    \State $AlivePrc \gets AlivePrc / maxAlive$;\vspace{1mm}
    \State $Power \gets A[currNodeDeed].power$;\Comment{Either capacity or actual}\vspace{1mm}
    \State $NPI \gets exp(Power) * AlivePrc$;\Comment{the Node Power Index is calculated}\vspace{1mm}
    \State $FP \gets 0$\vspace{1mm}
    \For{\textit{X in A }}\vspace{1mm}
        \State $currAlive \gets A[X].totalAliveTime$;\vspace{1mm}
        \State $O_x \gets currAlive/maxAlive$;\vspace{1mm}
        \State $I_x \gets exp(A[X].power) * O_x$;\vspace{1mm}
        \State $FP \gets FP + I_x$;\vspace{1mm}
    \EndFor\vspace{1mm}
    \State $R_{N_e} \gets NPI / FP$\vspace{1mm}
    \State \textbf{return} $R_{N_e}$
\end{algorithmic}
\end{algorithm}

\end{document}